\documentclass[runningheads]{llncs}
\usepackage[T1]{fontenc}
\usepackage{dsfont}
\usepackage{amsmath}
\usepackage{graphicx}
\usepackage{subfigure}
\begin{document}
\title{Univariate Channel Fusion for Multivariate Time Series Classification}
%
%
\author{Fernando Moro\inst{1} \and
Vinicius M. A. Souza\inst{1,2}}
\authorrunning{Moro and Souza}
%
\institute{Pontifícia Universidade Católica do Paraná (PUCPR), Brazil \and
Graduate Program in Informatics (PPGIa), Brazil\\
\email{{fernando.moro@pucpr.edu.br}, {vinicius.mourao@pucpr.br}}}
\maketitle              
\begin{abstract}
Multivariate time series classification (MTSC) plays a crucial role in various domains, including biomedical signal analysis and motion monitoring. However, existing approaches, particularly deep learning models, often require high computational resources, making them unsuitable for real-time applications or deployment on low-cost hardware, such as IoT devices and wearable systems. In this paper, we propose the Univariate Channel Fusion (UCF) method to deal with MTSC efficiently. UCF transforms multivariate time series into a univariate representation through simple channel fusion strategies such as the mean, median, or dynamic time warping barycenter. This transformation enables the use of any classifier originally designed for univariate time series, providing a flexible and computationally lightweight alternative to complex models. We evaluate UCF in five case studies covering diverse application domains, including chemical monitoring, brain-computer interfaces, and human activity analysis. The results demonstrate that UCF often outperforms baseline methods and state-of-the-art algorithms tailored for MTSC, while achieving substantial gains in computational efficiency, being particularly effective in problems with high inter-channel correlation.

\end{abstract}

\section{Introduction}

Biological signals, such as electroencephalography (EEG)~\cite{birbaumer1999spelling}, electrocardiography (ECG)~\cite{goldberger2000physiobank}, and electrocorticography (ECoG)~\cite{lal2004methods}, as well as signals collected for motion monitoring~\cite{souza2018asphalt}, naturally generate time series. Time series are sequences of real-valued observations recorded at equally spaced intervals and in temporal order~\cite{souza2025visemble}. A common requirement in applications that process such time series is the generation of rapid responses, either in real time or at least close to the arrival of new observations. For instance, in brain activity classification or motion monitoring, timely processing is crucial for responsive brain-computer interfaces, wearable health devices, or real-time movement analysis systems.

Consequently, the use of complex and computationally expensive models, such as deep learning architectures~\cite{ismail2019deep}, for time series classification can be challenging. Particularly when these models need to be deployed on low-cost hardware with limited memory and processing capabilities~\cite{da2021open}. In addition, time series are often collected at high sampling rates by sensors (e.g., 500~Hz), resulting in high-dimensional data~\cite{souza2021efficient}. This challenge is further amplified in the multivariate setting, where multiple variables (or channels) are observed simultaneously~\cite{dhariyal2020examination}. For instance, ECoG records brain activity from multiple sensors distributed across cortical regions, requiring models to process dozens or even hundreds of high-dimensional time series efficiently.

Approaches that transform multivariate series into univariate, for example by concatenating all channels or by training independent univariate models per channel, represent possible solutions~\cite{ruiz2021great}. However, even these seemingly simple strategies can become computationally expensive and memory-intensive when dealing with series comprising hundreds of variables.

In this paper, we propose the \textit{Univariate Channel Fusion} (UCF) method for efficiently addressing multivariate time series classification (MTSC). UCF transforms multivariate time series into a univariate representation by fusing channels through simple strategies such as the mean, median, or dynamic time warping barycenter~\cite{petitjean2011global}. In addition to being straightforward to implement and computationally efficient, the method is flexible, allowing the use of any classifier designed for univariate time series, such as ROCKET~\cite{dempster2020rocket} or QUANT~\cite{dempster2024quant}.


The main contributions of this paper are as follows:
\begin{itemize}
 \item We introduce UCF, a lightweight and classifier-agnostic strategy that fuses all channels of a multivariate time series into a single univariate series;
 \item We demonstrate that UCF achieves competitive accuracy and high efficiency on five real-world case studies from diverse domains, including brain-computer interfaces, chemical monitoring, and human activity analysis using EEG, ECG, near infrared spectroscopy, and ECoG data;
 \item We provide an analysis that elucidates the strengths and limitations of UCF, offering practical guidance on when channel fusion is effective.
\end{itemize}

Section~\ref{sec:preliminaries} introduces the main definitions and notation for multivariate time series. Section~\ref{sec:relatedwork} reviews relevant literature and Section~\ref{sec:proposal} presents the proposed UCF method. Section~\ref{sec:results} reports the experimental evaluation, followed by a discussion in Section~\ref{sec:discussion} and concluding remarks in Section~\ref{sec:conclusion}.

\vspace{-0.1cm}
\section{Preliminaries}\label{sec:preliminaries}
\vspace{-0.3cm}
A time series $X = \{x_1, x_2, \ldots, x_n\} \in \mathds{R}^{n \times l}$ is an ordered sequence of $n \in \mathds{N}$ observations collected over time. Each observation $x_t \in \mathds{R}^{l}$ can be represented as a vector $x_t = (x_{t,1}, x_{t,2}, \ldots, x_{t,l})$, 
where $l$ denotes the number of \textit{variables} or \textit{channels} measured at time $t$. In this work, the term \textit{channel} is used interchangeably with \textit{variable} to denote multivariate time series.

When $l = 1$, the series is said to be \textit{univariate}, meaning that only a single channel is measured over time (e.g., the daily temperature of a city). When $l > 1$, the series is \textit{multivariate}, consisting of simultaneous measurements of multiple channels (e.g., the $x$, $y$, and $z$ components of a three-axis accelerometer sensor).

\begin{figure}[htb]
    \centering
    \includegraphics[width=0.63\linewidth]{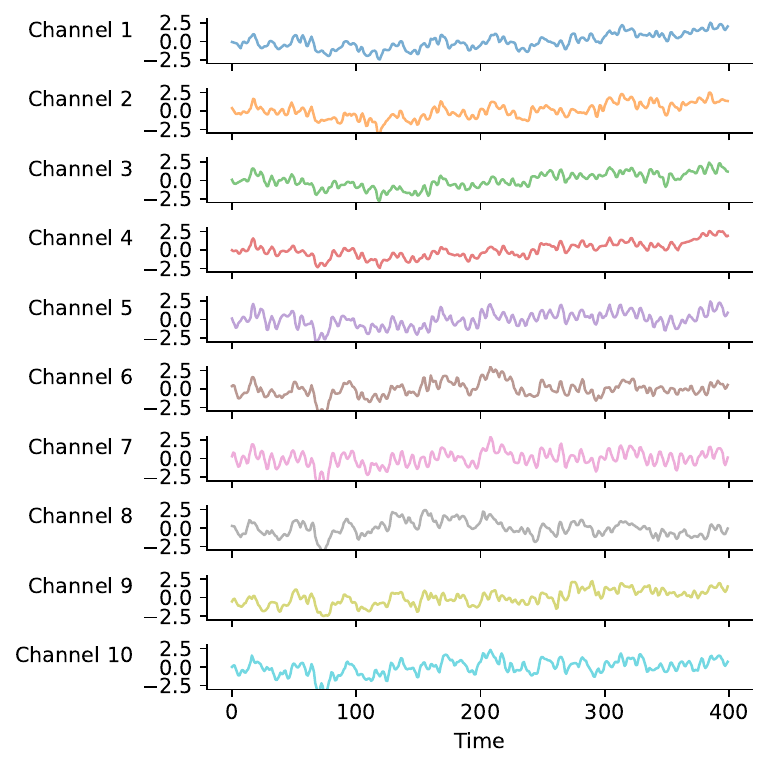}
    \vspace{-0.4cm}
    \caption{A multivariate time series with ten channels obtained by MEG activity.}
    \label{fig:multivariate_example}
\end{figure}

Fig.~\ref{fig:multivariate_example} illustrates a multivariate time series composed of magnetoencephalography (MEG) signals recorded while a subject performs a forward wrist movement using a joystick. In this example, the number of channels is $l=10$, and the series length is $n=400$. All signals are normalized to have a mean close to zero~\cite{lima2023large}. Further details of this example are provided in Section~\ref{subsec:handmovement}.
\vspace{-0.5cm}

\begin{figure}[htb]
    \centering 
    \includegraphics[trim={0 0.6cm 0 0},clip, width=0.83\linewidth]{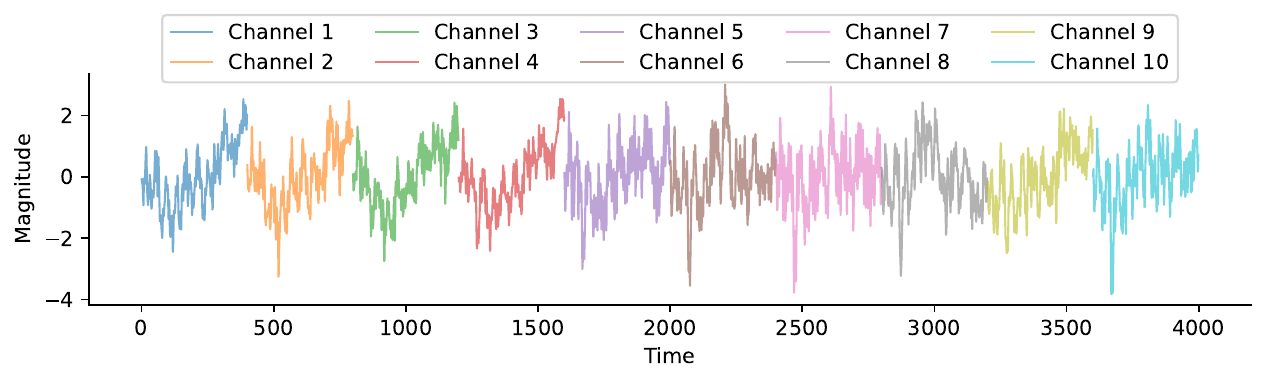}
    \vspace{-0.3cm}
    \caption{Multivariate time series converted into univariate via channel concatenation.}
    \label{fig:example_concatenation}
\end{figure}

\vspace{-0.3cm}
In some cases, it is convenient to convert a multivariate time series into a univariate representation. A straightforward strategy is to concatenate the channels into a single sequence, as illustrated in Fig.~\ref{fig:example_concatenation}, derived from the previous example. In this case, the 10 channels, each with 400 observations, were combined into a single univariate series with 4,000 observations, i.e., $l=1$ and $n=4,000$.
\vspace{-0.7cm}

\vspace{-0.5cm}
We focus on the task of \textit{multivariate time series classification}. Thus, each multivariate time series $X_i \in \mathds{R}^{n \times l}$ is associated with a single class label $y_i \in Y$, where $Y$ is a predefined set of class labels. Given a dataset of $m$ multivariate time series $\mathcal{X} = \{X_1, X_2, \ldots, X_m\}$ with corresponding labels $\mathcal{Y} = \{y_1, y_2, \ldots, y_m\}$, the goal of MTSC is to learn a classifier $f: \mathds{R}^{n \times l} \rightarrow Y$ that predicts the class label $y$ of previously unseen multivariate time series.  For the example shown in Fig.~\ref{fig:multivariate_example}, the set of possible classes is $Y = \{\textit{backward}, \textit{forward}, \textit{left}, \textit{right}\}$, and the illustrated instance $X_i$ corresponds to the class $y_i = \textit{forward}$.

\section{Related Work}\label{sec:relatedwork}

Algorithms for MTSC can be broadly categorized into two groups: (i) methods tailored for the multivariate setting, and 
(ii) methods adapted to use univariate models. The first group includes approaches explicitly designed to exploit the relationships among channels. These methods often integrate mechanisms to capture inter-channel correlations or perform joint feature extraction across dimensions. The second group encompasses strategies that adapt univariate models to multivariate data without explicitly modeling inter-channel dependencies. 

Representative algorithms from the first group include WEASEL+MUSE~\cite{Schafer2018muse}, TapNet~\cite{zhang2020tapnet}, and Nearest Neighbor with Dependent Dynamic Time Warping ($DTW_D$)~\cite{shokoohi2017generalizing}. WEASEL+MUSE builds a multivariate bag-of-patterns representation by extracting symbolic features from sliding windows in each channel and concatenating them into a joint feature vector, which is subsequently weighted and classified using a linear model. TapNet is a neural architecture that jointly learns temporal and variable-level representations through convolutional layers and a channel-attention mechanism, complemented by prototype-based embedding learning. In contrast, $DTW_D$ extends classic DTW to the multivariate setting by computing local alignment costs using multidimensional Euclidean distances, thus preserving inter-channel dependencies within a 1NN classifier.

The adaptation of univariate models to multivariate time series data can be achieved in several ways, such as concatenating all channels into a single long vector~\cite{ruiz2021great}, selecting the most discriminative channel (i.e., channel selection)~\cite{rushbrooke2025channel}, or training independent classifiers for each channel and combining their outputs through ensemble strategies such as majority voting~\cite{campagner2024ensemble}. Although simple, these approaches offer flexibility by allowing the reuse of well-established univariate models, often achieving competitive performance at a lower computational cost.

Another strategy that does not explicitly model inter-channel dependencies is the Independent Dynamic Time Warping ($DTW_I$)~\cite{shokoohi2017generalizing}. In this method, the classical DTW distance is applied independently to each channel of a multivariate time series. 
The resulting channel-wise distances are then aggregated, typically by averaging or summing, to produce a single similarity score between two multivariate series, which is used within a 1NN classifier.

Beyond these categories, some works explore forms of channel fusion that are conceptually related to ours. For example, Bai et al. introduced the Correlative Channel-Aware Fusion (C$^2$AF) network for multi-view time series classification, where multiple views correspond to different representations extracted from the same univariate series~\cite{bai2021correlative}. C$^2$AF first learns view-specific temporal features and then constructs a graph-based correlation matrix that models both intra- and inter-view relationships. This correlation structure is subsequently used by a learnable fusion module, implemented with convolutional blocks, to adaptively combine information across views. Although C$^2$AF also performs a form of correlation-driven fusion, it operates at the representation level and preserves multiple learned views, whereas our method performs direct signal-level fusion, converting a multivariate time series into a single univariate series.

\section{Proposed Method: Univariate Channel Fusion}\label{sec:proposal}
In this section, we introduce our proposed method, \textit{Univariate Channel Fusion} (UCF), which transforms multivariate time series into a univariate representation through channel fusion. This transformation enables the use of any univariate time series model on data that originally contain multiple correlated channels.

Our approach offers two main advantages. First, there is a much larger variety of models designed for univariate time series than for multivariate ones, allowing users to select the method that best suits their needs. Second, by reducing the number of channels through fusion, both training and inference times are significantly decreased, as we experimentally show.

We consider multivariate time series $X = \{x_1, x_2, \ldots, x_n\} \in \mathds{R}^{n \times l}$, where each observation $x_t = (x_{t,1}, x_{t,2}, \ldots, x_{t,l})$ represents the values of $l$ channels at time step $t$. The goal of our proposal is to transform the multivariate series into a univariate series $z = \{z_1, z_2, \ldots, z_n\} \in \mathds{R}^n$ by aggregating the values across channels into a single series. For pointwise aggregation methods (i.e., those that operate on each time step independently), this is achieved by applying an aggregation function $g: \mathds{R}^{l} \rightarrow \mathds{R}$ to each observation:

\[
z_t = g(x_t), \quad \text{for } t = 1, \ldots, n.
\]

In this work, we investigate three aggregation functions, although other choices are possible:

\begin{itemize}
    \item \textbf{Mean:} the average across channels at each time step, 
    $g(x_t) = \frac{1}{l} \sum_{j=1}^{l} x_{t,j}$.
    
    \item \textbf{Median:} the median value across channels at each time step, 
    $g(x_t) = \text{median}(x_{t,1}, \ldots, x_{t,l})$.
    
    \item \textbf{Barycenter:} the univariate series obtained by jointly aggregating all channels using DTW Barycenter Averaging (DBA)~\cite{petitjean2011global}. 
    A barycenter is a representative series that minimizes the sum of DTW distances to all input channels, effectively capturing the central tendency while preserving temporal alignments. 
    Formally, for the multivariate series $X$, let $X^{(j)} = \{x_{1,j}, \ldots, x_{n,j}\}$ denote the $j$-th channel. The fused series $z = \{z_1, \ldots, z_n\}$ is computed as
    $z = \text{DBA}(X^{(1)}, \ldots, X^{(l)})$,
    using the subgradient descent implementation, based on the stochastic subgradient mean algorithm, which iteratively estimates a barycenter sequence representing all channels.
\end{itemize}

\begin{figure}[htb]
    \centering
    \includegraphics[width=\linewidth]{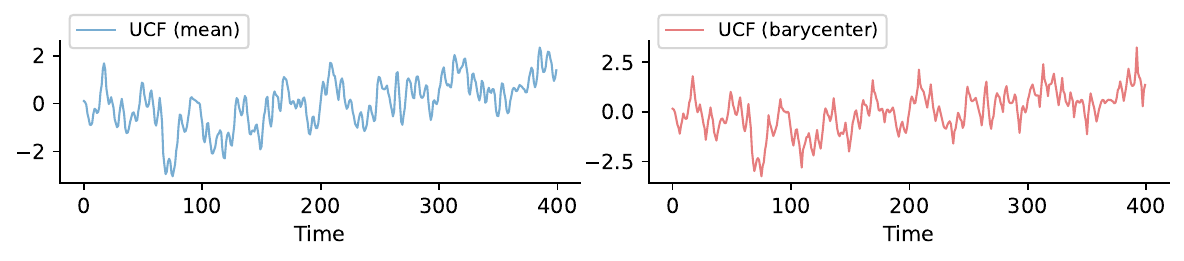}
    \vspace{-0.6cm}
    \caption{Example of a multivariate time series with ten channels converted into univariate representations through mean (\textit{left}) and barycenter (\textit{right}) aggregation functions.}
    \label{fig:UFC_example}
\end{figure}

Given the ten-channel multivariate example previously discussed in Section~\ref{sec:preliminaries}, Fig.~\ref{fig:UFC_example} illustrates the resulting univariate time series obtained through mean and barycenter aggregation functions to fuse multiple channels. As expected, the results are similar, with subtle differences. However, these differences can still impact the classifier's performance, as will be discussed later.

Beyond the previously discussed advantages, our approach also reduces the series length compared with simple methods such as channel concatenation. For example, UCF represents the series of Fig.~\ref{fig:multivariate_example} with 400 observations, whereas concatenation requires 4,000. This difference grows linearly with the number of channels, potentially becoming a bottleneck for the concatenation approach.

After transforming the original multivariate series into a univariate one through channel fusion, the resulting representation can be used as input to any model designed for univariate time series, or even to general-purpose machine learning algorithms that are not specifically tailored for temporal data. In our evaluation, we employ the QUANT classifier~\cite{dempster2024quant}.

QUANT is an interval-based classifier tailored for time series that extracts informative features from the series intervals. The algorithm computes quantiles over a fixed set of dyadic intervals and, in addition, applies three transformations: first-order differences, second-order differences, and a Fourier transform. For each set of intervals, the window is shifted by half the interval length to increase the number of extracted features. The features obtained from the original series, intervals, and transformations are then used to train an ensemble. 

\section{Experiments and Case Studies}\label{sec:results}
We demonstrate the capabilities of our proposal in five case studies from various domains, including brain-computer interfaces, chemical monitoring, and activity analysis. Firstly, we present a description of our experimental setup.

All codes, additional results (e.g., channel correlation heatmaps, UCF results with other aggregation functions), and data used in this article are publicly available on the supporting website of this work\footnote{\url{https://sites.google.com/view/ucf-mtsc/}}.

\subsection{Compared Methods}
We compare our approach against four baselines: (i) channel concatenation, (ii) channel ensemble by majority voting, (iii) $DTW_D$, and (iv) $DTW_I$. For a fair comparison, the first two baselines employ the same classifier used in our approach (i.e., QUANT), whereas $DTW_D$ and $DTW_I$ use a 1NN classifier, as in their original formulations. We further compare our approach against two state-of-the-art MTSC methods: (v) WEASEL+MUSE and (vi) TapNet.

These methods were selected as they represent naive, classical, and strong modern approaches, respectively. Although deep learning methods such as ResNet and InceptionTime have also achieved remarkable results in MTSC, we focus here on lightweight baselines to highlight the potential of channel fusion.


\subsection{Settings}
All algorithms were implemented in Python, and the experiments were conducted on a workstation with the following specifications: AMD Ryzen 9 7950X 16-core CPU, 128 GB of RAM, and an NVIDIA GeForce RTX 4090 GPU with 24 GB of VRAM. Among the methods, only TapNet required a GPU accelerator. All methods were evaluated using default hyperparameters and without tuning.

\subsection{Case Study 1: Cursor Control via EEG Signals}
Patients with severe motor impairments, such as those in a locked-in state, are unable to communicate through conventional means. To address this challenge, Birbaumer et al.~\cite{birbaumer1999spelling} introduced a brain–computer interface (BCI) that enabled the control of a computer cursor using slow cortical potentials (SCPs) measured from EEG signals. Thus, individuals who are completely paralyzed can learn to modulate their cortical activity, enabling them to move a cursor on a screen and select letters, thereby facilitating basic communication.

\begin{figure}[htb]
    \centering
    \includegraphics[width=0.8\linewidth]{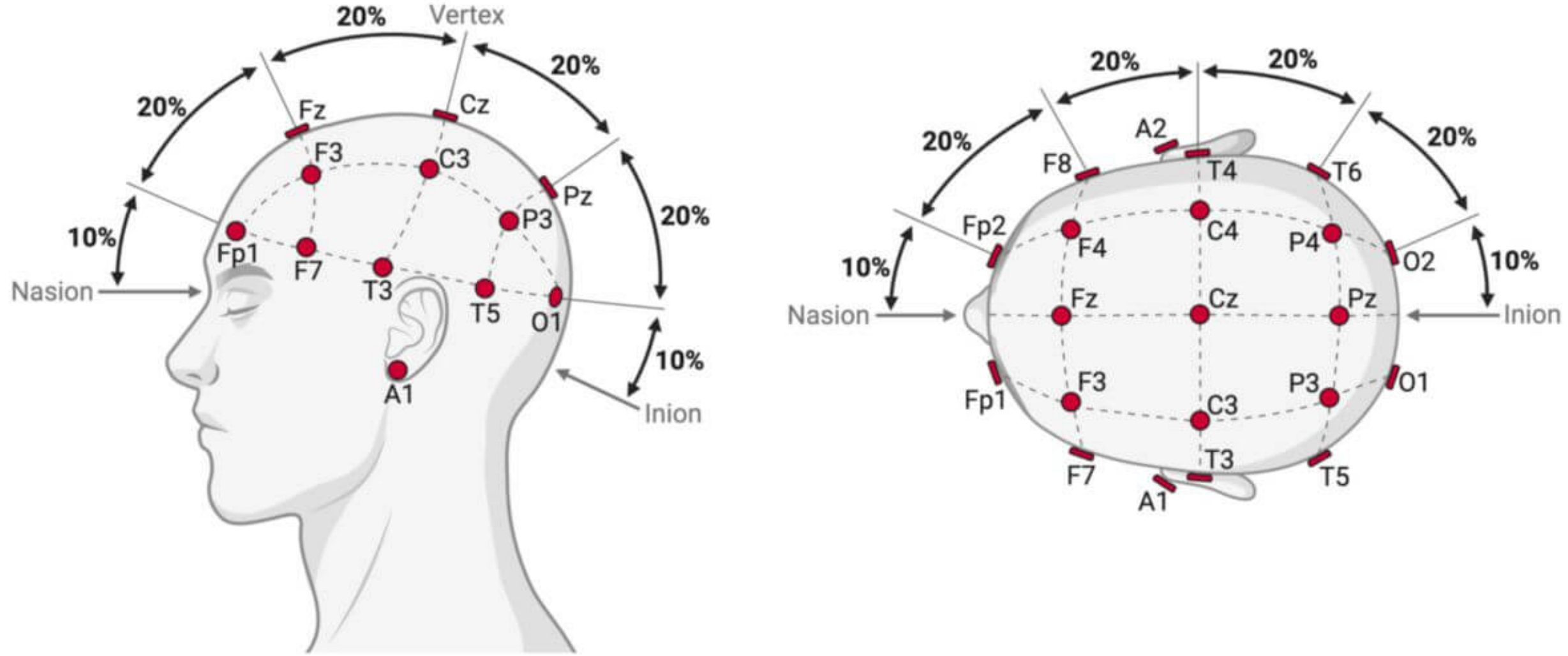}
    \vspace{-0.2cm}
    \caption{Standard 10-20 EEG electrode positioning system.}
    \label{fig:SCP}
\end{figure}

We evaluate our algorithm on EEG data collected from a healthy subject during a cursor control task, originally recorded in the study by Birbaumer et al.~\cite{birbaumer1999spelling}. The subject was instructed to move a cursor up or down by voluntarily generating cortical positivity or negativity, respectively, while receiving real-time visual feedback of their EEG activity recorded from the central scalp position (Cz) with reference electrodes placed on the mastoid bones behind the ears. EEG signals were sampled at 256~Hz, and the recording length of 3.5 s resulted in a time series with 896 observations per channel for each trial. Six channels were recorded according to the 10–20 system (Fig.~\ref{fig:SCP}), in which Channel 1: A1; Channel 2: A2; Channel 3: 2 cm anterior to C3; Channel 4: 2 cm posterior to C3; Channel 5: 2 cm anterior to C4; and Channel 6: 2 cm posterior to C4.
\vspace{-0.5cm}

Each 3.5s trial is labeled according to the intended cursor direction (upward or downward), defining a binary classification problem. The dataset comprises 268 labeled trials recorded over two sessions on different days, which form the training set. The test set contains 293 trials recorded on the second day. The goal is to predict the correct class for each trial in the test set.

The results in Table~\ref{tab:SelfRegulationSCP1} show that our proposed method, UCF, in its barycenter, median, and mean variants, achieved the top three accuracy scores, outperforming baseline methods such as DTW$_D$ and DTW$_I$, as well as both state-of-the-art algorithms (WEASEL+MUSE and TapNet). The computational efficiency of UCF is also notable. Using the barycenter, UCF requires less than 3 minutes to transform multivariate data into univariate series, train the model, and classify all test examples. Using the median, the accuracy remains high (87.37\%), with a processing time of under 2 seconds. In contrast, TapNet requires approximately 23 minutes, DTW$_D$ takes more than a day, and DTW$_I$ takes 53 minutes.
\vspace{-0.3cm}
\begin{table}[htb]
\caption{Results for cursor control via EEG signals.}
\scriptsize
\centering
\begin{tabular}{rcc}
\hline
\textbf{Algorithm} & \textbf{Accuracy} & \textbf{Time} \\ \hline
    WEASEL+MUSE & 79.18 & 3.32 min  \\ 
    TapNet & 79.18 & 22.86 min  \\
    QUANT (concatenation) & 80.54 & 9.27 s\\ 
    QUANT (ensemble) & 79.86 & 8.49 s \\ 
    DTW$_D$ & 77.47 & 30.64 h \\
    DTW$_I$ & 76.45 & 53.37 min  \\  \hline 
    UCF (mean) & 83.95 & 1.41 s\\ 
    UCF (median) & 87.37  & 1.53 s \\ 
    UCF (barycenter) & 87.71 & 2.98 min  \\ 
\hline
\end{tabular}
\label{tab:SelfRegulationSCP1}
\end{table}

\vspace{-1cm}
\subsection{Case Study 2: Motion Artifact Effects in ECG Signals} \label{subsec:standwalkjump}
\begin{figure}[htb]
    \centering
    \includegraphics[trim={0 0 0 3.5cm},clip,width=0.75\linewidth]{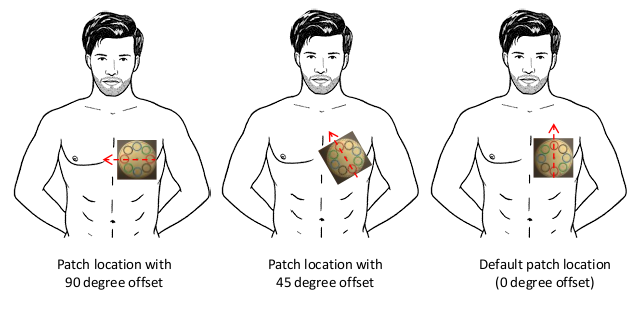}
    \vspace{-0.4cm}
    \caption{Four pairs of electrodes embedded in a single patch placed on the subject’s left chest. The patch was positioned at multiple orientations to simulate realistic variations in electrode placement during movement~\cite{goldberger2000physiobank}.}
    \label{fig:StandWalkJump}
\end{figure}

In this case study, we investigate the effects of motion artifacts on ECG signals using the dataset from Goldberger et al.~\cite{goldberger2000physiobank}. Short-duration recordings (5 s each) were obtained from a healthy 25-year-old male performing three physical activities: standing, walking, and jumping. Four pairs of electrodes embedded in a single patch captured separate ECG signals at 500 Hz, resulting in a 4-channel time series with 2,500 observations. During data collection, the patch was positioned at multiple orientations to simulate realistic variations in electrode placement during movement, as illustrated in Fig.~\ref{fig:StandWalkJump}. This dataset provides a controlled yet realistic scenario for evaluating classification algorithms designed to detect motion artifacts in ECG signals, which is crucial for practical applications such as wearable health monitoring, fitness tracking, and remote cardiac diagnostics.
\vspace{-0.3cm}

The results for this 3-class problem are shown in Table~\ref{tab:StandWalkJump}. The best result is achieved by UCF using the barycenter, with an accuracy of 66.66\%, followed by UCF with the median fusion strategy (60\%). Both outperform rival methods tailored for MTSC, such as WEASEL+MUSE and TapNet, which reached 40\% accuracy, as well as the QUANT classifier, whether by concatenating all channels into a single one or by training a separate classifier for each channel (ensemble). In addition to higher accuracy, UCF variants also show remarkable computational efficiency: while TapNet requires more than 5 minutes and DTW$_D$ over 30 minutes of processing, all UCF configurations complete within seconds.

\begin{table}[htb]
\caption{Results for motion effects in ECG signals.}
\scriptsize
\centering
\begin{tabular}{rcc}
\hline
\textbf{Algorithm} & \textbf{Accuracy} & \textbf{Time} \\ \hline
    WEASEL+MUSE & 40.00 & 5.70 s \\ 
    TapNet & 40.00 & 5.41 min  \\
    QUANT (concatenation) & 46.66 & 0.87 s \\ 
    QUANT (ensemble) & 40.00 & 1.29 s \\ 
    DTW$_D$ & 26.66 & 33.05 min  \\
    DTW$_I$ & 40.00 & 39.18 s \\  \hline 
    UCF (mean) & 46.66 & 0.32 s \\ 
    UCF (median) & 60.00 & 0.32 s \\ 
    UCF (barycenter) & 66.66 & 11.68 s \\ 
\hline
\end{tabular}
\label{tab:StandWalkJump}
\end{table}

\vspace{-0.7cm}
\subsection{Case Study 3: Ethanol Concentration Detection from Vibrational Spectroscopy}

Alcoholic beverages are often subject to adulteration and counterfeiting. Beyond economic impacts such as tax evasion and loss of brand integrity, these practices can pose serious health risks. For instance, in some countries, methanol is used as an adulterant, leading to severe poisoning, blindness, or even death~\cite{ilbeigi2025rapid}.

In this case study, we train a classifier to detect the ethanol concentration in alcoholic beverages. We use the dataset provided by Large et al.~\cite{large2018detecting}, which was acquired through non-invasive near-infrared spectroscopy from 44 distinct whisky bottles. The task consists of distinguishing four ethanol concentration levels (35\%, 38\%, 40\%, and 45\%) based on their vibrational spectral signatures. 

Vibrational spectroscopy in the near-infrared (NIR) measures light absorption associated with molecular vibrations, enabling the detection of subtle chemical differences such as ethanol concentration. Each spectrum was recorded from 226 to 1101.5 nm at 0.5 nm intervals, yielding time series with 1,751 observations per channel. The dataset comprises three spectral channels covering distinct wavelength regions, forming a 3-channel multivariate time series used for classifying ethanol concentrations. In this study, 261 multivariate series are used for training and 263 for testing.

The results are presented in Table~\ref{tab:EthanolConcentration}. The best performances were achieved by our proposal, UCF, with the mean (68.44\%) and barycenter (68.06\%) fusion strategies, followed by QUANT, which employed an ensemble voting scheme across channels (65.77\%). The lowest accuracy was obtained by TapNet (27.37\%), which also required considerably more processing time (54.9~min), whereas UCF completed the task in approximately 4~s.

\begin{table}[htb]
\caption{Results for the ethanol concentration detection.}
    \scriptsize
    \centering
    \begin{tabular}{rcc}
    \hline
         \textbf{Algorithm} & \textbf{Accuracy} & \textbf{Time} \\ \hline
         WEASEL+MUSE & 53.23 & 1.9 min \\ 
         TapNet & 27.37 & 54.9 min  \\
         QUANT (concatenation) & 62.35 & 14.9 s \\ 
         QUANT (ensemble) & 65.77 & 13.2 s \\ 
         DTW$_D$ & 34.22 & 100.4 h  \\
         DTW$_I$ & 30.79 & 1.5 h \\  \hline 
         UCF (mean) & 68.44 & 4.3 s \\ 
         UCF (median) & 63.11 & 4.5 s \\ 
         UCF (barycenter) & 68.06 & 2.6 min \\ 
         \hline
    \end{tabular}
    \label{tab:EthanolConcentration}
\end{table}

\vspace{-1cm}
\subsection{Case Study 4: Hand Movement Direction from MEG Signals}\label{subsec:handmovement}

\begin{figure}[htb]
    \centering
    \includegraphics[width=0.75\linewidth]{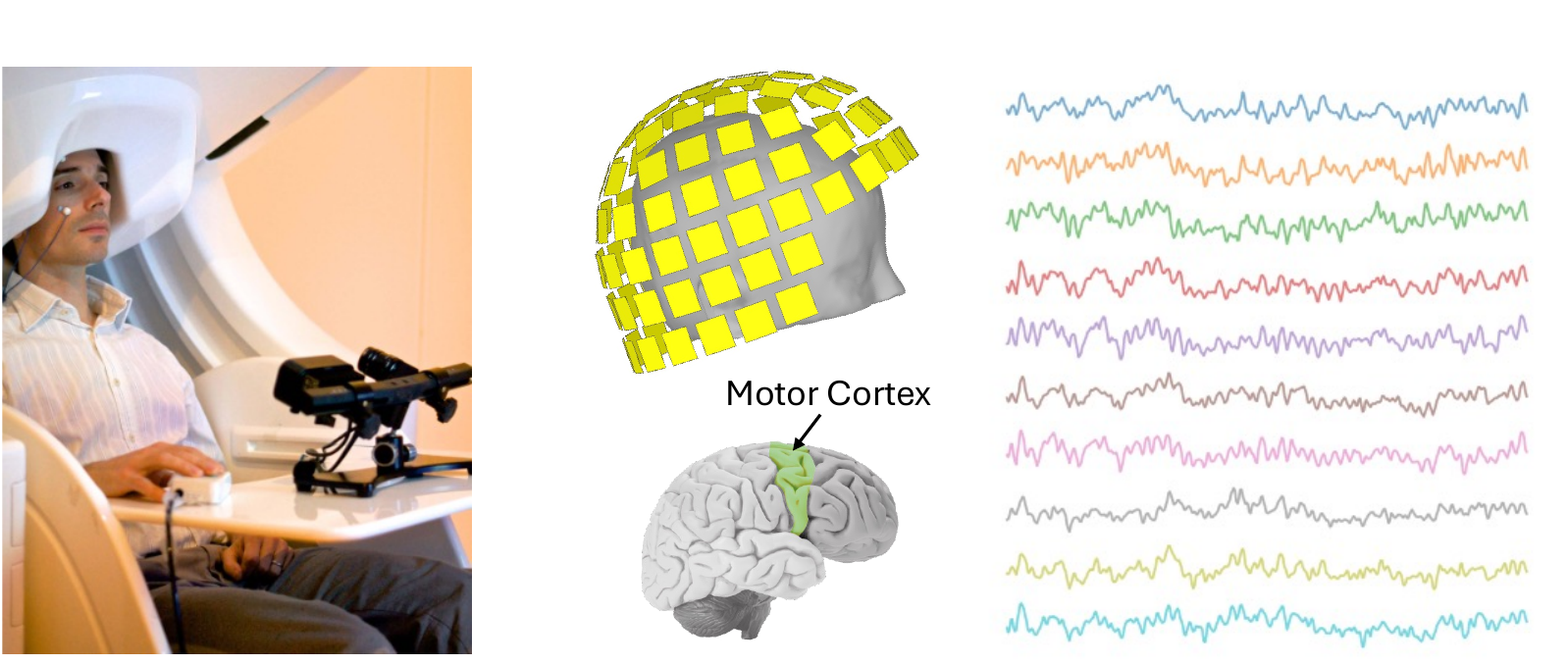}
    \caption{MEG recording session showing the sensor placement around the head and the corresponding multichannel signal data.}
    \label{fig:MEG_system}
\end{figure}

In this case study, we consider the dataset provided by the BCI Competition IV\footnote{\url{https://bbci.de/competition/iv/}}. The goal of this competition is to classify four wrist movements of subjects based on directionally modulated low-frequency magnetoencephalography (MEG) activity~\cite{tangermann2012review}. MEG is a non-invasive neuroimaging technique that measures brain activity with millisecond precision, allowing to localize the neural sources underlying movement-related processes. Fig.~\ref{fig:MEG_system} illustrates a MEG recording session. The left side of the figure shows the subject seated inside the MEG system~\cite{proudfoot2014magnetoencephalography}, while the center highlights the placement of the sensors and the location of the motor cortex. An example of data from 10 channels is shown on the right.
\vspace{-0.3cm}

Brain activity during wrist movements was recorded at 625~Hz from two healthy, right-handed subjects. The task consisted of moving a joystick from a central position toward one of four radially arranged targets at 90\textdegree ~intervals using only the right hand and wrist. Each trial was segmented to include 0.4~s before and 0.6~s after movement onset, band-pass filtered between 0.5 and 100~Hz, and resampled to 400~Hz, resulting in 400 time observations per series. The dataset comprises 160 training and 74 test multivariate series, each containing ten MEG channels positioned above the motor cortex. An example of the multivariate series for this problem after the fusion process is shown in Fig.~\ref{fig:UFC_example}.

The results for this four-class problem are shown in Table~\ref{tab:HandMoveDirection}. The best performance was achieved by UCF using median fusion (41.89\%), followed by barycenter (36.48\%) and mean (35.13\%). All UCF variants outperform competing methods. Among these, the strongest rivals were TapNet and the QUANT ensemble, both reaching 33.78\%. However, while TapNet and QUANT required 6.4~min and 8.4~s, respectively, UCF median and mean ran in under 1~s, and the barycenter strategy ran in 23.1~s.

\begin{table}[htb]
\caption{Results for the hand movement direction classification.}
    \centering
    \scriptsize
    \begin{tabular}{rcc}
    \hline
         \textbf{Algorithm} & \textbf{Accuracy} & \textbf{Time} \\ \hline
         WEASEL+MUSE & 25.67 & 33.7 s \\ 
         TapNet & 33.78 & 6.4 min \\ 
         QUANT (concatenation) & 31.08 & 6.5 s \\ 
         QUANT (ensemble) & 33.78 & 8.4 s \\ 
         DTW$_D$ & 22.97 & 56.1 min \\ 
         DTW$_I$ & 31.08 & 2.7 min \\ \hline 
         UCF (mean) & 35.13 & 0.8 s \\ 
         UCF (median) & 41.89 & 0.8 s \\ 
         UCF (barycenter) & 36.48 & 23.1 s \\ 
    \hline
    \end{tabular}
    \label{tab:HandMoveDirection}
\end{table}

\vspace{-0.5cm}
\subsection{Case Study 5: Motor Imagery Classification from ECoG}

In this case study, we consider a two-class classification problem from the BCI III competition\footnote{\url{https://bbci.de/competition/iii/desc\_I.html}}. During data collection, a subject imagined movements of either the left small finger or the tongue while electrocorticography (ECoG) was recorded~\cite{lal2004methods}. ECoG is an intracranial electrophysiological monitoring technique that records electrical activity directly from the cerebral cortex via electrodes placed on its exposed surface.

Electrical brain activity was captured using an $8 \times 8$ platinum ECoG grid, generating 64-channel time series. All recordings were sampled at 1000~Hz. Each trial started 0.5~s after a visual cue ended and lasted 3~s, resulting in 3000 time points per series. The training and test data were collected from the same subject performing the same task on two different days, approximately one week apart. The training set comprises 278 trials from the first session, while the test set contains 100 trials from the second session, which is particularly challenging, as the subject’s state in terms of motivation, fatigue, and other factors may differ between sessions, leading to variability in brain activity.

The results are presented in Table~\ref{tab:MotorImagery}. The highest accuracy was achieved by UCF using mean fusion (52\%), followed by QUANT with concatenation (51\%) and QUANT ensemble (50\%). Despite the similar accuracies, UCF required only 7.8~s, whereas the QUANT variants took 23.3~min and 10.9~min, respectively, a substantial difference. Tailored MTSC algorithms such as WEASEL+MUSE and TapNet performed below 50\% accuracy and required approximately 2~hours, highlighting the challenge of this problem.

\begin{table}[htb]
\caption{Results for the motor imagery classification.}
    \centering
    \scriptsize
    \begin{tabular}{rcc}
    \hline
         \textbf{Algorithm} & \textbf{Accuracy} & \textbf{Time} \\ \hline
         WEASEL+MUSE & 48 & 1.9 h \\ 
         TapNet & 47 & 2.1 h \\
         QUANT (concatenation) & 51 & 23.2 min \\ 
         QUANT (ensemble) & 50 & 10.9 min \\ 
         DTW$_D$ & 49 & 207.3 h \\ 
         DTW$_I$ & 47 & 39.0 h \\ \hline 
         UCF (mean) & 41 & 7.4 s \\ 
         UCF (median) & 52 & 7.8 s \\ 
         UCF (barycenter) & 43 & 1.7 h \\ 
    \hline
    \end{tabular}
    \label{tab:MotorImagery}
\end{table}

\vspace{-1cm}
\subsection{Statistical Analysis}
\begin{figure}[htb]
    \centering
    \subfigure[Accuracy]{
    \includegraphics[scale=0.46]{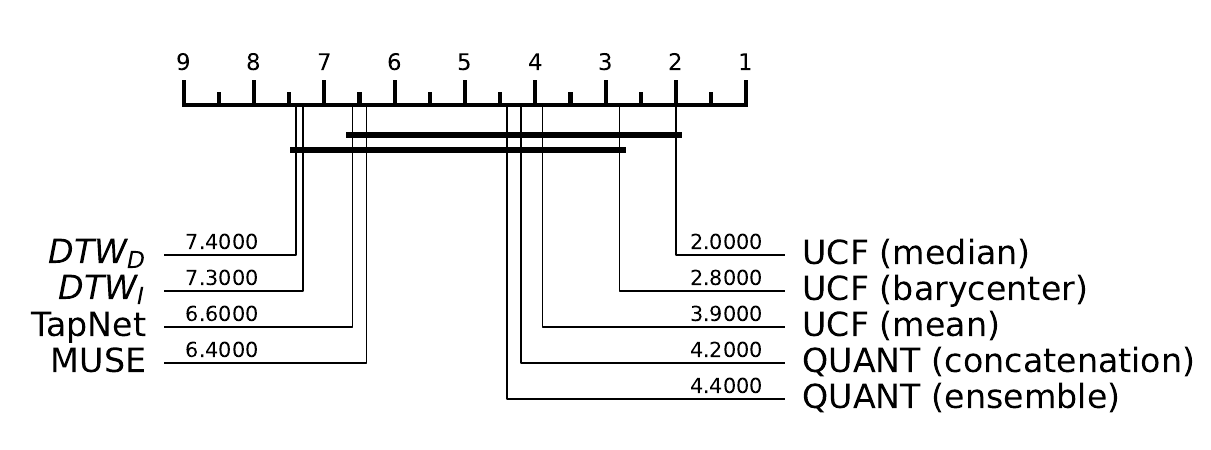}
   }
    \subfigure[Execution time]{
    \includegraphics[scale=0.46]{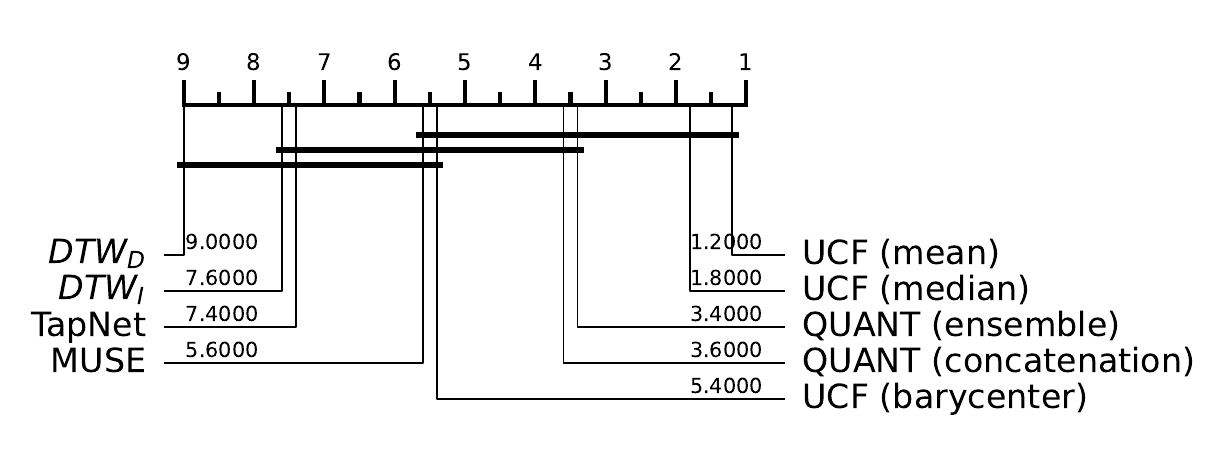}
    }
    \vspace{-0.2cm}
    \caption{Critical difference diagrams comparing the algorithms.}
    \label{fig:critical_difference}
\end{figure}

For the statistical comparison of both accuracy and computational efficiency, we generated critical difference diagrams (CDD) using the Nemenyi test with a significance level of $\alpha = 0.1$. This non-parametric test is well-suited for comparing multiple algorithms over a small number of datasets. The resulting diagrams are shown in Fig.~\ref{fig:critical_difference}.
\vspace{-0.5cm}

As shown in Fig.~\ref{fig:critical_difference}-(a), all UCF variants rank in the top positions in the accuracy CDD, with statistically significant differences only with DTW$_D$ and DTW$_I$. Regarding execution time in Fig.~\ref{fig:critical_difference}-(b), UCF (mean) and UCF (median) occupy the first two ranks, differing significantly from DTW$_D$, DTW$_I$, and TapNet. However, the statistical comparison is constrained by the limited number of datasets.

\section{Strengths and Limitations of UCF}\label{sec:discussion}
Despite its efficiency, UCF has limitations. Because the fusion step compresses multiple channels into a single univariate representation, some information loss is unavoidable. This loss may negatively impact performance in problems where inter-channel dependencies are critical to discrimination, particularly when channels are highly heterogeneous. 

\begin{figure}[htb]
    \centering
    \subfigure[StandWalkJump]{
    \includegraphics[scale=0.45]{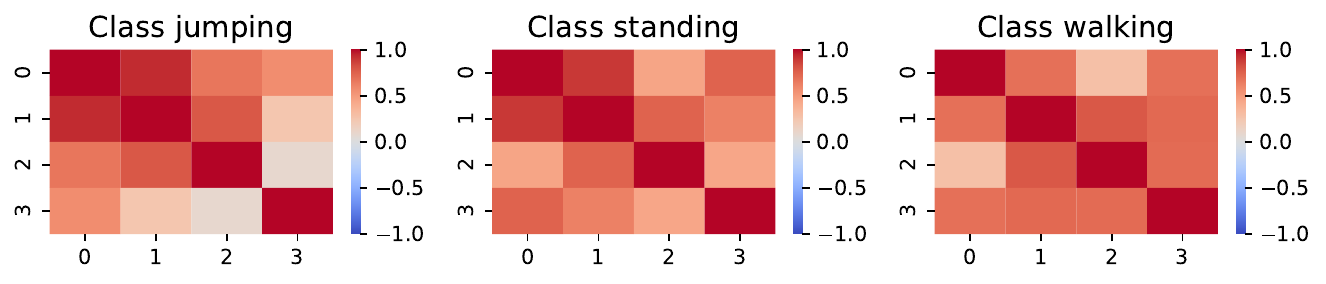}
   }
    \subfigure[BasicMotions]{
    \includegraphics[scale=0.4]{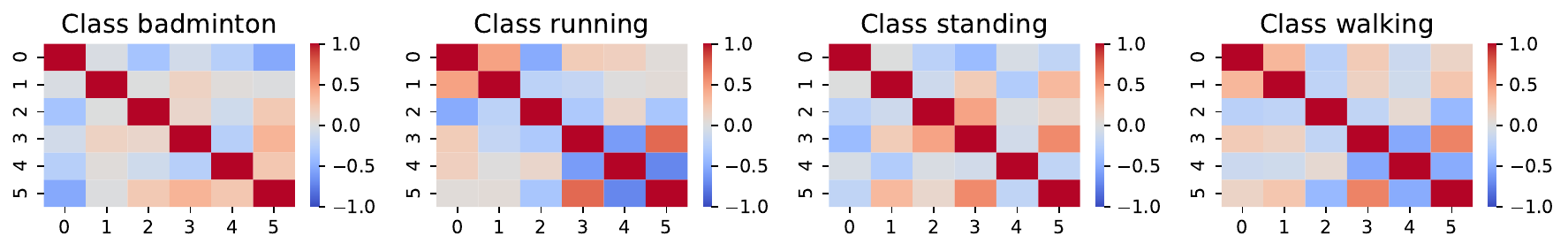}
    }
    \vspace{-0.4cm}
    \caption{Channel-wise correlation heatmaps for two activity recognition problems.}
    \label{fig:heatmaps}
\end{figure}

To better understand the applicability of UCF, we examine two similar activity-recognition problems in which the method succeeds in one case but fails in the other. We analyze these contrasting outcomes by inspecting the channel-correlation structure of each dataset. Using Pearson correlation, we generated channel-wise heatmaps to characterize the similarity across channels. Fig.~\ref{fig:heatmaps} presents these heatmaps separately for each class, highlighting how differences in inter-channel relationships help explain the divergent performance of UCF.

\vspace{-0.5cm}
\vspace{-0.4cm}
The first example corresponds to Case Study 2 (StandWalkJump), where the goal is to identify physical activities from ECG recordings. In this dataset, the four channels exhibit strong and consistent correlations within individual classes. Such redundancy suggests that much of the discriminative information is shared across channels, allowing UCF to fuse them into a single series without significant loss of class-relevant structure.

The second example corresponds to the BasicMotions dataset~\cite{bagnall2018uea}, which contains triaxial accelerometer and gyroscope signals recorded from a smartwatch during four physical activities. In this case, the correlation heatmaps reveal weak and highly variable dependencies across most of the six channels, reflecting differences mainly in the sensor modality. In this scenario, UCF tends to underperform because it collapses heterogeneous channels into a single univariate signal, which hides modality-specific information that is essential for discrimination.

These observations suggest that UCF is most effective when channels are highly correlated, such as when they originate from the same modality, and less suitable when channels provide complementary but non-redundant information. Heatmaps for all case studies are available on the supporting website.

\section{Conclusion}\label{sec:conclusion}

This paper introduced Univariate Channel Fusion (UCF), a simple yet effective method that converts multivariate time series into a univariate representation through channel fusion, enabling the use of any model designed for univariate time series classification. The method's simplicity and low computational cost make it particularly attractive for resource-constrained hardware such as IoT devices and wearable systems.

We demonstrated the effectiveness of UCF across five real-world case studies. The results show that UCF achieves superior accuracy compared to traditional baselines (DTW$_D$ and DTW$_I$), deep learning architectures (TapNet), and specialized algorithms for MTSC (WEASEL+MUSE), while maintaining remarkable computational efficiency. Among the evaluated fusion strategies, we recommend using the median, which provides a balanced trade-off between predictive performance and efficiency. Although the barycenter fusion achieved comparable accuracy, it was the most computationally demanding approach.

We recommend using UCF in MTSC problems where the channels originate from the same modality and exhibit high inter-channel correlation. In future work, we plan to investigate fusion strategies that operate on subsets of highly correlated channels rather than collapsing all channels into a single series.


%
%
%
 \bibliographystyle{splncs04}
 \bibliography{refs}

@inproceedings{large2018detecting,
  title={Detecting forged alcohol non-invasively through vibrational spectroscopy and machine learning},
  author={Large, James and Kemsley, E Kate and Wellner, Nikolaus and Goodall, Ian and Bagnall, Anthony},
  booktitle={Pacific-Asia conference on knowledge discovery and data mining},
  pages={298--309},
  year={2018},
  organization={Springer}
}

@article{tangermann2012review,
  title={Review of the BCI competition IV},
  author={Tangermann, Michael and M{\"u}ller, Klaus-Robert and Aertsen, Ad and Birbaumer, Niels and Braun, Christoph and Brunner, Clemens and Leeb, Robert and Mehring, Carsten and Miller, Kai J and M{\"u}ller-Putz, Gernot R and others},
  journal={Frontiers in neuroscience},
  volume={6},
  pages={55},
  year={2012},
  publisher={Frontiers Research Foundation}
}

@article{lal2004methods,
  title={Methods towards invasive human brain computer interfaces},
  author={Lal, Thomas and Hinterberger, Thilo and Widman, Guido and Schr{\"o}der, Michael and Hill, N and Rosenstiel, Wolfgang and Elger, Christian and Birbaumer, Niels and Sch{\"o}lkopf, Bernhard},
  journal={Advances in neural information processing systems},
  volume={17},
  year={2004}
}

@article{birbaumer1999spelling,
  title={A spelling device for the paralysed},
  author={Birbaumer, Niels and Ghanayim, Nimr and Hinterberger, Thilo and Iversen, Iver and Kotchoubey, Boris and K{\"u}bler, Andrea and Perelmouter, Juri and Taub, Edward and Flor, Herta},
  journal={Nature},
  volume={398},
  number={6725},
  pages={297--298},
  year={1999},
  publisher={Nature Publishing Group UK London}
}

@article{goldberger2000physiobank,
  title={PhysioBank, PhysioToolkit, and PhysioNet: components of a new research resource for complex physiologic signals},
  author={Goldberger, Ary L and Amaral, Luis AN and Glass, Leon and Hausdorff, Jeffrey M and Ivanov, Plamen Ch and Mark, Roger G and Mietus, Joseph E and Moody, George B and Peng, Chung-Kang and Stanley, H Eugene},
  journal={circulation},
  volume={101},
  number={23},
  pages={e215--e220},
  year={2000},
  publisher={Lippincott Williams \& Wilkins}
}

@inproceedings{zhang2020tapnet,
  title={Tapnet: Multivariate time series classification with attentional prototypical network},
  author={Zhang, Xuchao and Gao, Yifeng and Lin, Jessica and Lu, Chang-Tien},
  booktitle={Proceedings of the AAAI conference on artificial intelligence},
  volume={34},
  number={04},
  pages={6845--6852},
  year={2020}
}

@inproceedings{Schafer2018muse,
  author    = {Patrick Sch{\"a}fer and Ulf Leser},
  title     = {Multivariate Time Series Classification with {WEASEL+MUSE}},
  booktitle = {Proceedings of the 3rd ECML/PKDD Workshop on Advanced Analytics and Learning on Temporal Data (AALTD)},
  pages = {1-11},  
year      = {2018}
}

@article{shokoohi2017generalizing,
  title={Generalizing DTW to the multi-dimensional case requires an adaptive approach},
  author={Shokoohi-Yekta, Mohammad and Hu, Bing and Jin, Hongxia and Wang, Jun and Keogh, Eamonn},
  journal={Data mining and knowledge discovery},
  volume={31},
  number={1},
  pages={1--31},
  year={2017},
  publisher={Springer}
}

@article{petitjean2011global,
  title={A global averaging method for dynamic time warping, with applications to clustering},
  author={Petitjean, Fran{\c{c}}ois and Ketterlin, Alain and Gan{\c{c}}arski, Pierre},
  journal={Pattern recognition},
  volume={44},
  number={3},
  pages={678--693},
  year={2011},
  publisher={Elsevier}
}

@article{dempster2024quant,
  title={Quant: A minimalist interval method for time series classification},
  author={Dempster, Angus and Schmidt, Daniel F and Webb, Geoffrey I},
  journal={Data Mining and Knowledge Discovery},
  volume={38},
  number={4},
  pages={2377--2402},
  year={2024},
  publisher={Springer}
}

@inproceedings{rushbrooke2025channel,
  title={Channel Selection and Creation Algorithms for Electroencephalography Classification with HIVE-COTE},
  author={Rushbrooke, Aiden and Middlehurst, Matthew and Sami, Saber and Bagnall, Anthony},
  booktitle={HAIS},
  pages={328--339},
  year={2025},
  organization={Springer}
}

@article{ruiz2021great,
  title={The great multivariate time series classification bake off: a review and experimental evaluation of recent algorithmic advances},
  author={Ruiz, Alejandro Pasos and Flynn, Michael and Large, James and Middlehurst, Matthew and Bagnall, Anthony},
  journal={Data mining and knowledge discovery},
  volume={35},
  number={2},
  pages={401--449},
  year={2021},
  publisher={Springer}
}

@article{campagner2024ensemble,
  title={Ensemble predictors: Possibilistic combination of conformal predictors for multivariate time series classification},
  author={Campagner, Andrea and Barandas, Marilia and Folgado, Duarte and Gamboa, Hugo and Cabitza, Federico},
  journal={IEEE Transactions on Pattern Analysis and Machine Intelligence},
  volume={46},
  number={11},
  pages={7205--7216},
  year={2024},
  publisher={IEEE}
}

@article{lima2023large,
  title={A large comparison of normalization methods on time series},
  author={Lima, Felipe Tomazelli and Souza, V M A},
  journal={Big Data Research},
  volume={34},
  pages={100407},
  year={2023},
  publisher={Elsevier}
}

@article{ilbeigi2025rapid,
  title={Rapid and Direct Determination of Methanol in Alcoholic Beverages by Ion Mobility Spectrometry},
  author={Ilbeigi, Vahideh and Valadbeigi, Younes and Matejcik, Stefan},
  journal={Analytical Chemistry},
  volume={97},
  number={33},
  pages={18265--18272},
  year={2025},
  publisher={ACS Publications}
}

@article{da2021open,
  title={An open-source tool for classification models in resource-constrained hardware},
  author={Silva, Lucas Tsutsui and Souza, V M A and Batista, Gustavo EAPA},
  journal={IEEE Sensors Journal},
  volume={22},
  number={1},
  pages={544--554},
  year={2021},
  publisher={IEEE}
}

@article{bagnall2018uea,
  title={The UEA multivariate time series classification archive, 2018},
  author={Bagnall, Anthony and Dau, Hoang Anh and Lines, Jason and Flynn, Michael and Large, James and Bostrom, Aaron and Southam, Paul and Keogh, Eamonn},
  journal={arXiv preprint arXiv:1811.00075},
  year={2018}
}

@inproceedings{dhariyal2020examination,
  title={An examination of the state-of-the-art for multivariate time series classification},
  author={Dhariyal, Bhaskar and Le Nguyen, Thach and Gsponer, Severin and Ifrim, Georgiana},
  booktitle={ICDM workshops},
  pages={243--250},
  year={2020},
  organization={IEEE}
}

@article{souza2021efficient,
  title={Efficient unsupervised drift detector for fast and high-dimensional data streams},
  author={Souza, V M A and Parmezan, Antonio RS and Chowdhury, Farhan A and Mueen, Abdullah},
  journal={Knowledge and Information Systems},
  volume={63},
  number={6},
  pages={1497--1527},
  year={2021},
  publisher={Springer}
}

@article{souza2018asphalt,
  title={Asphalt pavement classification using smartphone accelerometer and complexity invariant distance},
  author={Souza, V M A},
  journal={Engineering Applications of Artificial Intelligence},
  volume={74},
  pages={198--211},
  year={2018},
  publisher={Elsevier}
}

@article{souza2025visemble,
  title={Visemble: A fast ensemble approach for time series classification with multiple visual representations},
  author={Souza, Vinicius M A and Veiga, Patrickerson S and Ribeiro, Andre G R},
  journal={Knowledge-Based Systems},
  volume={309},
  pages={112864},
  year={2025},
  publisher={Elsevier}
}

@article{ismail2019deep,
  title={Deep learning for time series classification: a review},
  author={Ismail Fawaz, Hassan and Forestier, Germain and Weber, Jonathan and Idoumghar, Lhassane and Muller, Pierre-Alain},
  journal={Data mining and knowledge discovery},
  volume={33},
  number={4},
  pages={917--963},
  year={2019},
  publisher={Springer}
}

@article{proudfoot2014magnetoencephalography,
  title={Magnetoencephalography},
  author={Proudfoot, Malcolm and Woolrich, Mark W and Nobre, Anna C and Turner, Martin R},
  journal={Practical neurology},
  volume={14},
  number={5},
  pages={336--343},
  year={2014},
  publisher={BMJ Publishing Group Ltd}
}

@article{dempster2020rocket,
  title={ROCKET: exceptionally fast and accurate time series classification using random convolutional kernels},
  author={Dempster, Angus and Petitjean, Fran{\c{c}}ois and Webb, Geoffrey I},
  journal={Data Mining and Knowledge Discovery},
  volume={34},
  number={5},
  pages={1454--1495},
  year={2020},
  publisher={Springer}
}

@inproceedings{bai2021correlative,
  title={Correlative channel-aware fusion for multi-view time series classification},
  author={Bai, Yue and Wang, Lichen and Tao, Zhiqiang and Li, Sheng and Fu, Yun},
  booktitle={Proceedings of the AAAI conference on artificial intelligence},
  volume={35},
  number={8},
  pages={6714--6722},
  year={2021}
}

\end{document}